\UseRawInputEncoding
\documentclass[prl,twocolumn,superscriptaddress,floatfix,nopacs,nofootinbib,longbibliography]{revtex4-2}

\usepackage{graphicx,amsfonts,amssymb,amsmath,hyperref,enumerate}

\usepackage{algorithm}
\usepackage{algpseudocode}
\usepackage{pgfplots}
\usepackage{pgf}
\usepackage{lmodern}
\usepackage{import}
\usepackage{xr}
\usepackage{enumitem}
\usepackage{soul}
\usepackage[normalem]{ulem}
\usepackage{multirow}
\newif\ifhyper
\hypertrue
\ifhyper
\hypersetup{
   citecolor = {red},
   colorlinks = {true}, 
   urlcolor = {blue} 
}
\fi

\newcommand{\beq}{\begin{equation}}
\newcommand{\eeq}{\end{equation}}
\newcommand{\beqa}{\begin{eqnarray}}
\newcommand{\eeqa}{\end{eqnarray}}

\newcommand{\comment}[1]{}

\def\Longarrow{\protect\@lra}
\def\@lra{\relbar\joinrel\relbar\joinrel\relbar\joinrel\relbar\joinrel\rightarrow}

\begin{document} 

\title{CompactifAI: {Extreme} Compression of Large Language Models \\ {using} Quantum-Inspired Tensor Networks}

\author{Andrei Tomut}
\affiliation{Multiverse Computing, Parque Cientifico y Tecnol\'{o}gico de Gipuzkua, Paseo de Miram\'{o}n, 170 3$^{\,\circ}$ Planta, 20014 Donostia / San Sebasti\'{a}n, Spain}
\affiliation{Catalan Institute of Nanoscience and Nanotechnology (ICN2), CSIC and The Barcelona Institute of Science and
Technology, Campus UAB, 08193 Bellaterra, Catalonia, Spain.}

\author{Saeed S. Jahromi}
\affiliation{Donostia International Physics Center, Paseo Manuel de Lardizabal 4, E-20018 San Sebasti\'an, Spain}
\affiliation{Multiverse Computing, Parque Cientifico y Tecnol\'{o}gico de Gipuzkua, Paseo de Miram\'{o}n, 170 3$^{\,\circ}$ Planta, 20014 Donostia / San Sebasti\'{a}n, Spain}

\author{Abhijoy Sarkar} 
\affiliation{Multiverse Computing, Parque Cientifico y Tecnol\'{o}gico de Gipuzkua, Paseo de Miram\'{o}n, 170 3$^{\,\circ}$ Planta, 20014 Donostia / San Sebasti\'{a}n, Spain}

\author{Uygar Kurt} 
\affiliation{Multiverse Computing, Parque Cientifico y Tecnol\'{o}gico de Gipuzkua, Paseo de Miram\'{o}n, 170 3$^{\,\circ}$ Planta, 20014 Donostia / San Sebasti\'{a}n, Spain}

\author{Sukhbinder Singh}
\affiliation{Multiverse Computing, Centre for Social Innovation, 192 Spadina Avenue Suite 509, Toronto, ON M5T 2C2 Canada}

\author{Faysal Ishtiaq} 
\affiliation{Multiverse Computing, Parque Cientifico y Tecnol\'{o}gico de Gipuzkua, Paseo de Miram\'{o}n, 170 3$^{\,\circ}$ Planta, 20014 Donostia / San Sebasti\'{a}n, Spain}

\author{C\'esar Mu\~noz}
\affiliation{Multiverse Computing, Parque Cientifico y Tecnol\'{o}gico de Gipuzkua, Paseo de Miram\'{o}n, 170 3$^{\,\circ}$ Planta, 20014 Donostia / San Sebasti\'{a}n, Spain}

\author{Prabdeep Singh Bajaj}
\affiliation{Multiverse Computing, Parque Cientifico y Tecnol\'{o}gico de Gipuzkua, Paseo de Miram\'{o}n, 170 3$^{\,\circ}$ Planta, 20014 Donostia / San Sebasti\'{a}n, Spain}

\author{Ali Elborady}
\affiliation{Multiverse Computing, Parque Cientifico y Tecnol\'{o}gico de Gipuzkua, Paseo de Miram\'{o}n, 170 3$^{\,\circ}$ Planta, 20014 Donostia / San Sebasti\'{a}n, Spain}

\author{Gianni del Bimbo}
\affiliation{Multiverse Computing, Parque Cientifico y Tecnol\'{o}gico de Gipuzkua, Paseo de Miram\'{o}n, 170 3$^{\,\circ}$ Planta, 20014 Donostia / San Sebasti\'{a}n, Spain}

\author{Mehrazin Alizadeh}
\affiliation{Multiverse Computing, Centre for Social Innovation, 192 Spadina Avenue Suite 509, Toronto, ON M5T 2C2 Canada}

\author{David Montero}
\affiliation{Multiverse Computing, Parque Cientifico y Tecnol\'{o}gico de Gipuzkua, Paseo de Miram\'{o}n, 170 3$^{\,\circ}$ Planta, 20014 Donostia / San Sebasti\'{a}n, Spain}

\author{Pablo Mart\'in-Ramiro}
\affiliation{Multiverse Computing, Parque Cientifico y Tecnol\'{o}gico de Gipuzkua, Paseo de Miram\'{o}n, 170 3$^{\,\circ}$ Planta, 20014 Donostia / San Sebasti\'{a}n, Spain}

\author{Muhammad Ibrahim}
\affiliation{Multiverse Computing, Parque Cientifico y Tecnol\'{o}gico de Gipuzkua, Paseo de Miram\'{o}n, 170 3$^{\,\circ}$ Planta, 20014 Donostia / San Sebasti\'{a}n, Spain}

\author{Oussama Tahiri Alaoui}
\affiliation{Multiverse Computing, Parque Cientifico y Tecnol\'{o}gico de Gipuzkua, Paseo de Miram\'{o}n, 170 3$^{\,\circ}$ Planta, 20014 Donostia / San Sebasti\'{a}n, Spain}

\author{John Malcolm}
\affiliation{Multiverse Computing, Centre for Social Innovation, 192 Spadina Avenue Suite 509, Toronto, ON M5T 2C2 Canada}

\author{Samuel Mugel}
\affiliation{Multiverse Computing, Centre for Social Innovation, 192 Spadina Avenue Suite 509, Toronto, ON M5T 2C2 Canada}

\author{Rom\'{a}n Or\'{u}s}
\email{roman.orus@multiversecomputing.com}

\affiliation{Multiverse Computing, Parque Cientifico y Tecnol\'{o}gico de Gipuzkua, Paseo de Miram\'{o}n, 170 3$^{\,\circ}$ Planta, 20014 Donostia / San Sebasti\'{a}n, Spain}

\affiliation{Donostia International Physics Center, Paseo Manuel de Lardizabal 4, E-20018 San Sebasti\'an, Spain}

\affiliation{Ikerbasque Foundation for Science, Maria Diaz de Haro 3, E-48013 Bilbao, Spain}

\begin{abstract}
Large Language Models (LLMs) such as ChatGPT and LlaMA are advancing rapidly in generative Artificial Intelligence (AI), but their immense size poses significant challenges, such as huge training and inference costs, substantial energy demands, and limitations for on-site deployment. Traditional compression methods such as pruning, distillation, and low-rank approximation focus on reducing the effective number of neurons in the network, while quantization focuses on reducing the numerical precision of individual weights to reduce the model size while keeping the number of neurons fixed. While these compression methods have been relatively successful in practice, there is no compelling reason to believe that truncating the number of neurons is an optimal strategy. In this context, this paper introduces \textit{CompactifAI}, an innovative LLM compression approach using quantum-inspired Tensor Networks that focuses on the model's correlation space instead, allowing for a more controlled, refined and interpretable model compression. Our method is versatile and can be implemented with --- or on top of --- other compression techniques. As a benchmark, we demonstrate that a combination of \emph{CompactifAI} with quantization allows to reduce a 93\% the memory size of LlaMA-2 7B, reducing also 70\% the number of parameters, accelerating 50\% the training and 25\% the inference times of the model, and just with a small accuracy drop of 2\% - 3\%, going much beyond of what is achievable today by other compression techniques. Our methods also allow to perform a refined layer sensitivity profiling, showing that deeper layers tend to be more suitable for tensor network compression, which is compatible with recent observations on the ineffectiveness of those layers for LLM performance. Our results imply that standard LLMs are, in fact, heavily overparametrized, and do not need to be large at all. 

\end{abstract}

\maketitle

\emph{Introduction.-} The emergence of generative artificial intelligence (AI) has ushered in an era where computers can perform tasks that were unimaginable just a few years ago. A prime example of this advancement is found in Large Language Models (LLMs) \cite{llm}, which are based on the innovative ``transformer architecture." \cite{attention} The field of LLMs experienced a significant surge with the introduction of OpenAI's ChatGPT \cite{chatgpt}, showcasing an unprecedented level of human-machine interaction. Following this, several other models, such as Meta's LlaMA \cite{LlaMA} and Google's BERT \cite{bert}, were developed. Currently, LLMs are expected to be utilized not only in linguistic applications but also across various sectors, attracting substantial investments in this transformative technology. This development represents the most profound technological revolution since the inception of the internet.

However, LLMs are not without their challenges. The most significant issue is the energy consumption required for training these AI models. As noted by the CEO of OpenAI, training ChatGPT-3 incurred an estimated $100$ million dollars in electricity bills alone, and the costs for training such models are predicted to double every ten months \cite{economist}. Coupled with the exponentially growing demand for these systems, we face a daunting scenario: the development of these systems is currently unsustainable without significantly impacting the planet. The immense energy consumption of LLMs is untenable, compelling the need for greener, more efficient solutions. In this context, various compression techniques for LLMs have been suggested, with quantization \cite{Quantization}, distillation \cite{Distilling} , pruning \cite{Pruning}, and low-rank approximations \cite{Lora} being among the most prominent. However, these methods are quite brute-force --- they largely focus on truncating the effective number of neurons, even when the original model's accuracy is known to increase with size during training. Consequently, controlling and anticipating the compression error in these schemes is challenging, and their application has met with mixed success.

In this paper, we introduce \emph{CompactifAI}  \cite{compactifai}, a {novel LLM compression technique} based on quantum-inspired Tensor Networks (TNs) \cite{tn1, tn2}. This technique involves ``tensorizing" the self-attention (SA) and multi-layer perceptron (MLP) layers using a specific TN, which effectively truncates the correlations present in the model. The degree of truncation can be controlled via the bond dimension of the TN, \emph{enabling a significant reduction in the memory size and number of parameters} of the LLM model while maintaining accuracy. In practice, the compressed model requires less energy and memory, and operations such as training, retraining, and inference become more efficient. The ``tensorized" model is retrained using multi-GPU distributed training. Within this framework, we observed that the significant reduction in the number of model parameters by tensorization drastically reduces the GPU-CPU transfer time, consequently reducing the training and inference time in our benchmarks  by $50\%$ and $25\%$, respectively.  Hence, our tensorization approach is particularly well-suited for distributed training of LLMs. 
As we will demonstrate, a brief retraining period allows the accuracy of the compressed model to approach that of the original uncompressed version.

\begin{figure}
    \centering
    \includegraphics[scale = 0.4]{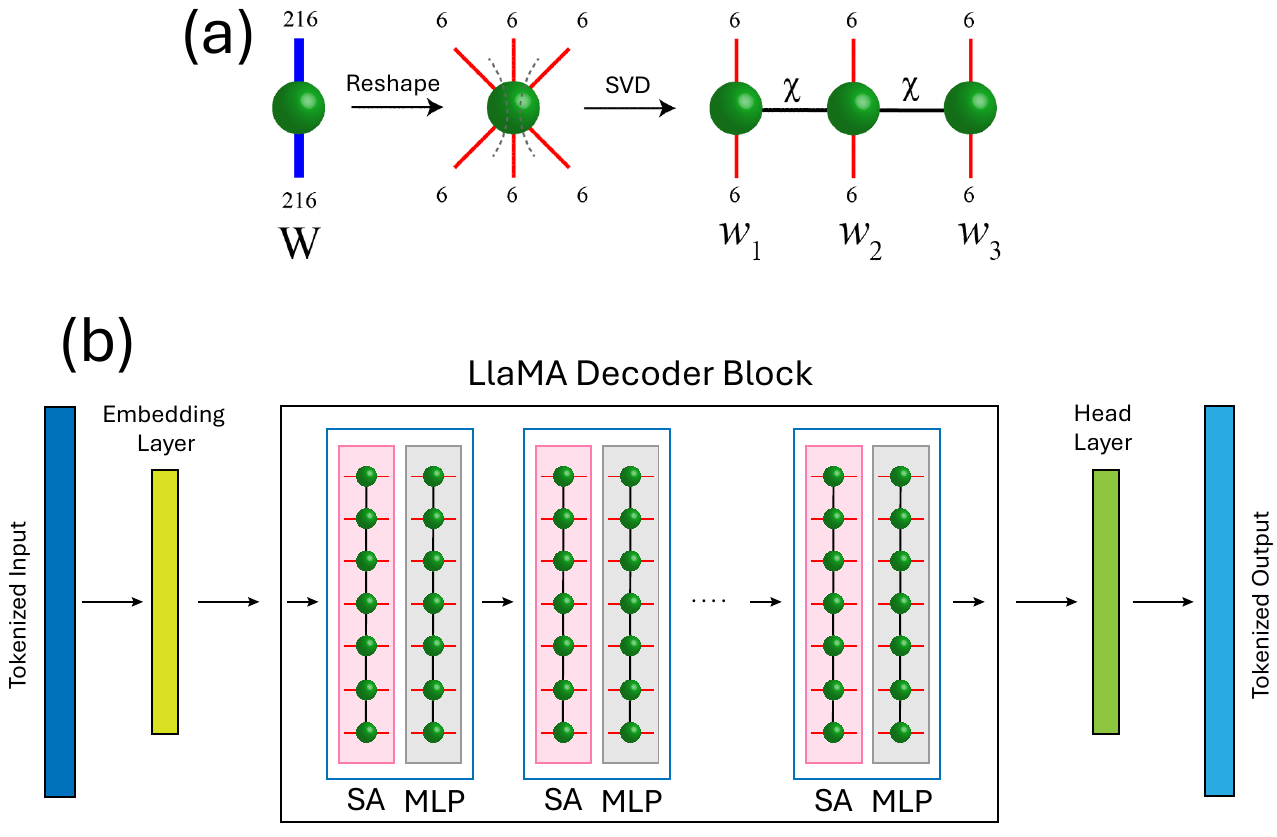}
    \caption{[Color online] (a) Example, in diagrammatic representation, of the decomposition of a weight matrix $W$ in terms of an MPO. The original matrix has $216 \times 216$ parameters. After reshaping the matrix indices followed by two sequential SVDs, the resulting tensor network has $2 \times 36 \chi + 36 \chi^2$ parameters, amounting to the sum of parameters of each tensor, with $\chi$ being the MPO bond dimension serving as a truncation parameter. In the diagrammatic representation of MPOs, circles represent individual tensors, lines indicate tensor indices and lines connecting circles represent contracted shared indices between tensors. (b) Schematic graphical representation of the tensorization mechanism for LLMs within the LlaMA model family (generalization to other LLM architectures is straightforward). Embedding and head layers customize the input and output for a given task, with the tokenized input/output comprising words and sentences. The Self Attention and Multi-layer Perceptron layers within the LlaMA decoder block are tensorized in such a way that the weight matrices of the corresponding neural networks are decomposed into appropriate Tensor Networks, in this case, MPOs with a bond dimension of $\chi$.}
    \label{Fig:architecture}
\end{figure}

\begin{table}
 \begin{center}
	\begin{tabular}{||c|c|c|c||}
 \hline
 Model & Size & Parameters &  Quantization \\ 
\hline
 Original & $27.1$ GB & $7$B   &   float-$32$\\
 $8$-bit  & $6.8$ GB  & $7$B   &   int-$8$ \\ 
 $4$-bit  & $3.4$ GB  & $7$B   &   int-$4$ \\ 
 $88\%$   & $4.1$ GB  & $2.1$B &   float-$16$\\ 
 $93\%$   & $2.1$ GB  & $2.1$B &   mixed \\
 \hline 
	\end{tabular}
 \end{center}
  \caption{Details of the models used in the benchmarks. The quantization in the $93\%$ compressed model is a mix of float-$16$ for the tensorized layers and int-$4$ quantization for the not-tensorized layers.}
 \label{tab1}
\end{table}

\begin{figure}
    \centering
    \includegraphics[width=\columnwidth]{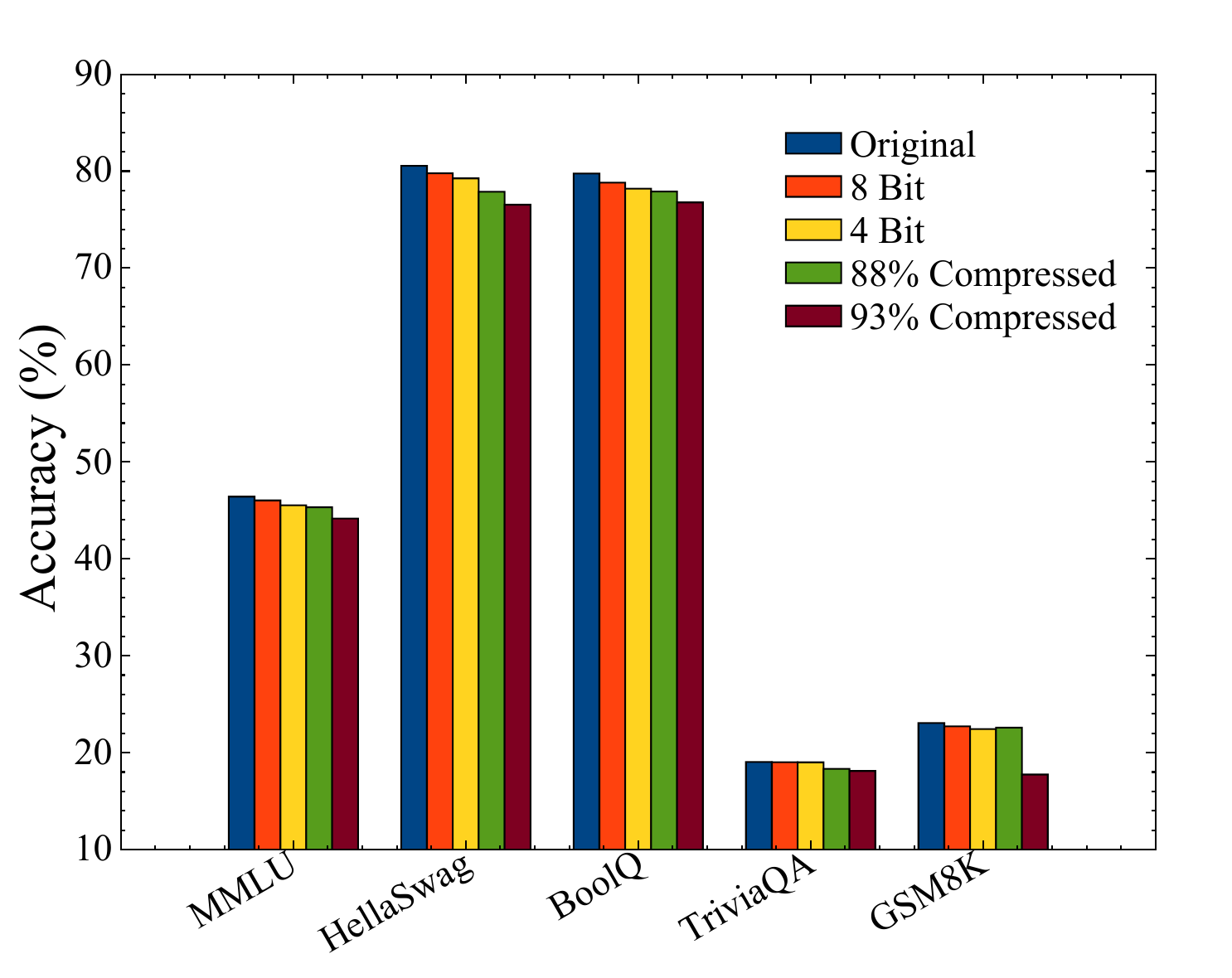}
    \caption{[Color online] Accuracies of the original and compressed models for the tasks related to Language Understanding (MMLU), Commonsense reasoning (HellaSwag), Reading comprehension (BoolQ), World knowledge (TriviaQA) and Math (GSM8K). The accuracy of the compressed models only deviates by $2\%$-$3\%$ compared to the original LlaMA 2 7B.}
    \label{Fig:accuracy}
\end{figure}

\bigskip 

\emph{Method.-} The compression method we propose is based on the efficient decomposition of weight matrices in neural networks into Tensor Networks, such as Matrix Product Operators (MPOs) and similar structures. This concept has been successfully implemented in deep learning architectures, as previously demonstrated \cite{tnn0, tnn1, tnn2, tnn3}, {but to the best of our knowledge, this work is the first application of this approach to compressing LLMs.} Specifically for Large Language Models (LLMs), our approach involves first, a layer sensitivity profiling (see Supplementary Information), that guides identifying the layers that are more tenable to correlation compression and then replacing their trainable weights with suitable TNs (in the present case, MPOs). The results of our layer sensitivity profiling are compatible with and in fact improve (and refine) upon recent observations that deeper layers tend to be ineffective in the performance of LLM models \cite{mit}. Without loss of generality, here we consider the case of the LLM architecture of LlaMA-2 chat models. As illustrated in Fig.\ref{Fig:architecture}, we substitute the weight matrices in the Self Attention (SA) and Multi-layer Perceptron (MLP) layers of the (pretrained) LlaMA decoder block with MPOs characterized by a bond dimension $\chi$. The process of determining the MPO involves executing sequential Singular Value Decompositions (SVDs) on the respective weight matrix, retaining the largest $\chi$ singular values at each SVD. This truncation in $\chi$ effectively truncates the correlations among model parameters within a given layer to only the most relevant ones necessary to describe the system, while discarding those that are irrelevant. The approach leads to a significant reduction in memory costs, as storing the truncated MPO, which incurs a polynomial cost, is far more efficient than storing the original weight matrix, which would require an exponential cost in the number of neurons within the layer. Furthermore, the bond dimension $\chi$ effectively controls the level of compression: a smaller $\chi$ results in more information being discarded, leading to greater compression but at the cost of reduced accuracy. The choice of TN architecture as well as the number of decomposed tensors for each layer can be considered as additional hyper-parameters for the compressed model. 

\begin{figure}
    \centering
    \includegraphics[width=\columnwidth]{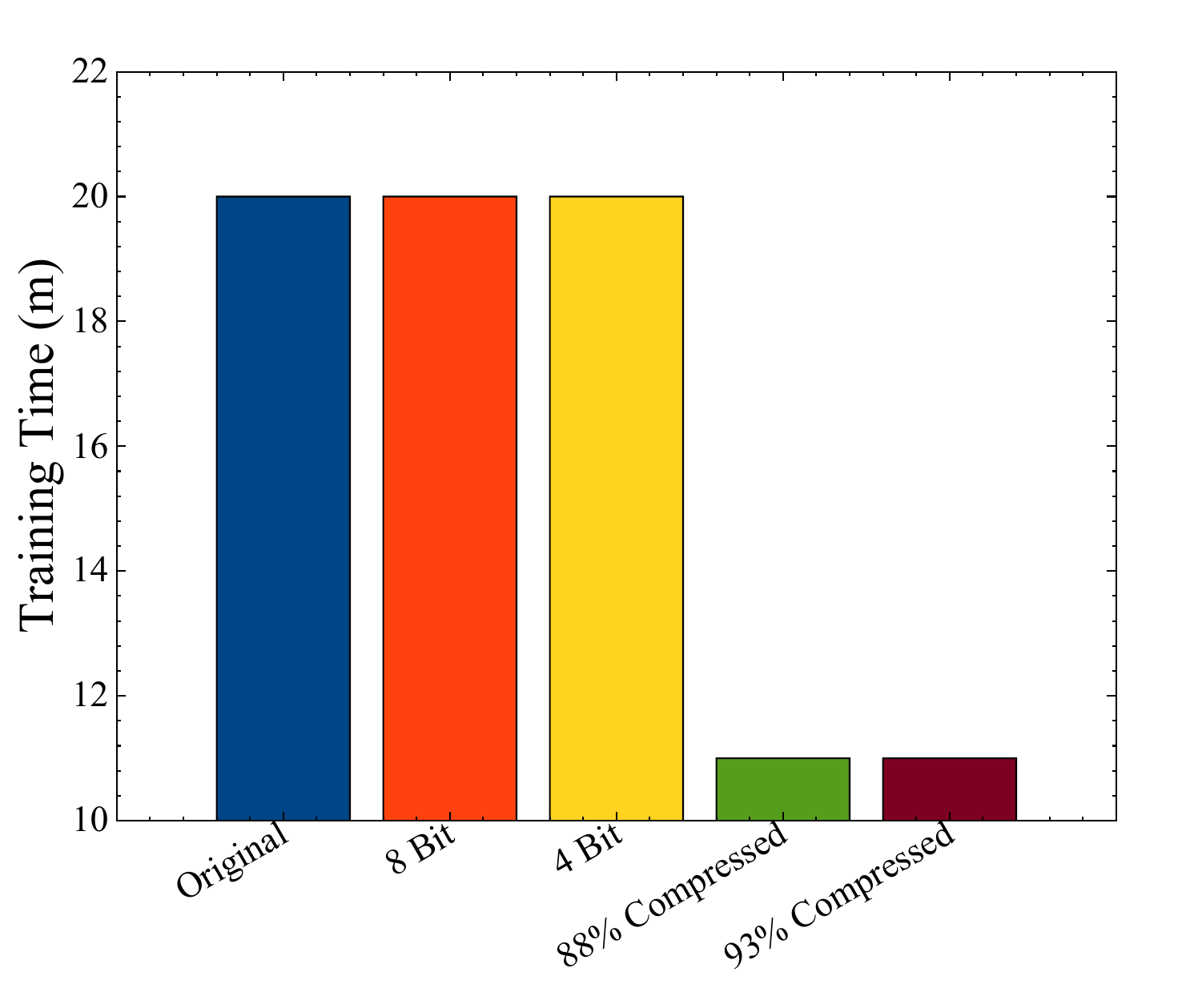}
    \caption{[Color online] Training time (in minutes) of the different models on the same amount of MMLU data used for healing the tensorized models. The tensorized models show $2$x speed up (i.e. half the time) with distributed training on eight A10g NVIDIA GPUs with respect to both the original and purely-quantized models.}
    \label{Fig:training-time}
\end{figure}

To ensure high accuracy in the compressed model, our method also includes a rapid retraining phase, dubbed as {\it healing}, following the determination of the truncated MPOs. This retraining is essential because the local, layer-by-layer truncation into MPOs -- akin to the so-called ``simple update" in TN algorithms \cite{simple} -- may not be optimal in general, in the sense that the effect of other layers are not explicitly taken into account when truncating the weight matrix of a specific layer. However, retraining the compressed structure is way more efficient than training the original uncompressed model due to the significantly smaller number of model parameters, which reduces the CPU-GPU transfer times in a distributed training setup. As we demonstrate below, after just a few retraining epochs of the compressed model, its accuracy closely approaches that of the original uncompressed model but at a fraction of the cost.

\begin{figure}
    \centering
    \includegraphics[width=\columnwidth]{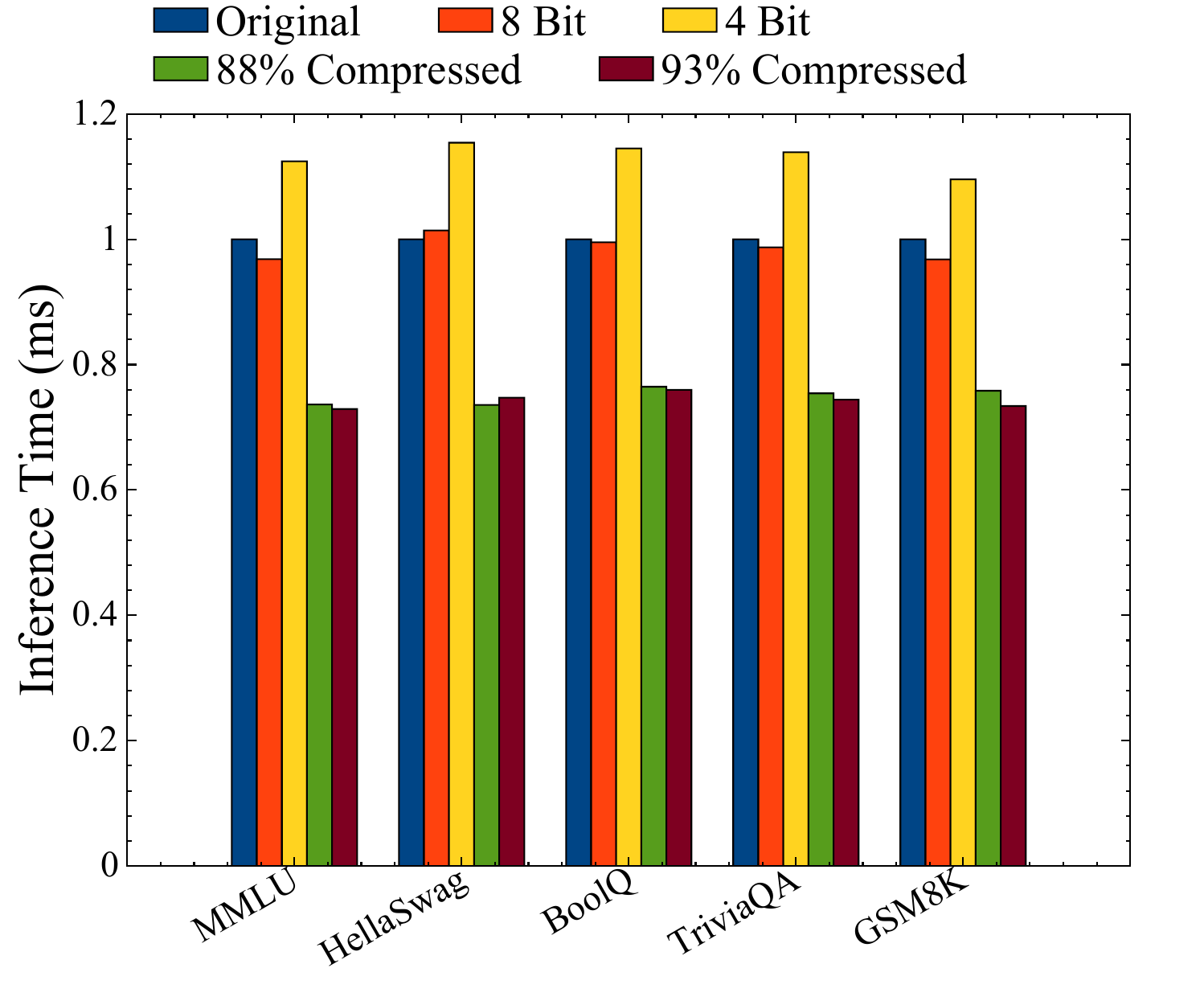}
    \caption{[Color online] Inference time (in milliseconds) of the different models for measuring the accuracies of the MMLU, HellaSwag, BoolQ, TriviaQA, and GSM8K tasks. The tensorized models are $25\%$ faster with distributed inference on eight A10g NVIDIA GPUs with respect to the original model. Notice that inference time of some quantized models is even higher than that of the original. The inference times are normalized by taking the time of the original model as a reference. }
    \label{Fig:inference-time}
\end{figure}

\bigskip 

\emph{Benchmark.-} To evaluate our method, we used it to compress the LlaMA-$2$ $7$B model. This model represents the ``smallest" within the ``large" category of LLMs in the open-source LlaMA series, developed by META. It encompasses $7$ billion parameters, has been pre-trained on over $2$ trillion tokens, offers a context length of $4096$, and has undergone fine-tuning with more than $1$ million human annotations. 

In order to benchmark the model we created several compressed versions of the LlaMA-$2$ $7$B models by a combination of tensor network compression and quantization. Details of the benchmarked models, such as their size in memory, number of parameters and their quantization are available in Table~\ref{tab1}. The $8$-bit and  $4$-bit quantized models were created from the original LlaMA model by using the \texttt{bitsandbytes} quantization library. Furthermore, the $88\%$ and $93\%$ compressed models were created by applying  \textit{CompactifAI} to the float-$16$ quantized version of the original LlaMA model. The parameter counts in Table~\ref{tab1} show that while the size of the models can be reduced both by tensor network compression and quantization, \emph{the tensorized model allows for more size reduction than quantization by significantly reducing the model parameters}. As we will show, unlike in quantization, this parameter reduction is the key feature that allows significant speed up in both training and inference. 

Let us further note that the tensorized models in Table~\ref{tab1} were healed after compression to recover the accuracy drop caused by parameter reduction. We used generic chat datasets such as \textit{Ultrachat}, \textit{Alpaca} and \textit{OpenHermess} to retrain the tensorized model, which was implemented on a single AWS EC2 instance with $8$ NVIDIA A10g GPU processors and distributed training. The healing process was performed for less than a single epoch on the aforementioned datasets. The $93\%$ compressed model was obtained by applying $4$-bit quantization to the not-tensorized layers of the $88\%$ compressed and healed model. 

Once the compressed models are healed, we benchmarked all the models of Table~\ref{tab1} in tasks related to language understanding (MMLU), commonsense reasoning (HellaSwag), reading comprehension (BoolQ), world knowledge (TriviaQA) and math (GSM8K). We further used the \texttt{LLM Evaluation Harness} \cite{harness} library to calculate the accuracies on these five tasks. 

\begin{table}
 \begin{center}
	\begin{tabular}{||c|c|c|c|c|c||}
 \hline 
 Task/Model & Original & $8$-bit &  $4$-bit  & $88\%$ & $93\%$ \\ 
 \hline
 MMLU      & $46.41$ & $46.03$ & $45.53$ & $45.32$ & $44.16$\\
 HellaSwag & $80.55$ & $79.77$ & $79.25$ & $77.87$ & $76.54$\\ 
 BoolQ     & $79.76$ & $78.81$ & $78.19$ & $77.90$ & $76.77$\\ 
 TriviaQ   & $19.03$ & $19.01$ & $19.00$ & $18.33$ & $18.10$\\ 
 GSM8K     & $23.05$ & $22.71$ & $22.44$ & $22.58$ & $17.74$\\   
 \hline 
	\end{tabular}
 \end{center}
  \caption{Accuracies of the models in Table~\ref{tab1} for the MMLU, HellaSwag, BoolQ, TriviaQA and GSM8K tasks.}
 \label{tab2}
\end{table}

Fig.~\ref{Fig:accuracy} shows the accuracy of the original and compressed models for the target tasks. While the accuracies of all compressed models are very close to those of the original $7$B LlaMA model and only deviate by $2\%-3\%$, the $88\%$ and $93\%$ compressed models reach the same level of accuracies with $70\%$ fewer parameters (just $2.1$ billion). This suggests that a substantial portion of the parameters in LLMs are redundant, and discarding even up to $70\%$ of the parameters does not degrade the model accuracy significantly. Such behavior suggests that, after all, \emph{large language models are heavily overparametrized, and they do not need to be large in practice}. For better clarity, we have further reported all the accuracies in Table~\ref{tab2} for the original and compressed models. Let us stress that the accuracies of the tensorized models are obtained by training only for one epoch during the healing, and better accuracies can be obtained with further fine-tuning of the tensorized models, sometimes even higher than that of the original, as we have seen in parallel tests with smaller models.    

On top of the accuracy results, another interesting observation that demonstrates the power of tensor network compression is the remarkable speedup that tensorized models manifest both during training and inference. In order to benchmark the training speed of all models in Table~\ref{tab1}, we trained the original and the quantized models on the same amount of data used for tensorized models, and measured their training times. Fig.~\ref{Fig:training-time} shows the training times of all models compared against each other. Tensorized models exhibit a remarkable $50\%$ acceleration (i.e. $2$x faster) compared to the original and purely quantized models. This significant speedup is attributed to the substantially smaller number of parameters in tensorized models, which are transferred much faster between the GPUs and CPUs during distributed training.  

 Next, we tested all models for inference time using distributed training with both data and model parallelization. Fig.~\ref{Fig:inference-time} shows the inference time (forward time of the model) for different models compared to the original model. Inference times are in milliseconds, and normalized with respect to that of the original model (dark blue bar in the plot). Let us point out that while tensorized models bring more than $25\%$ speed up in the inference, the $4$-bit quantized model, on the contrary, slows down the inference by $13\%$. This might be due to the fact that some quantized operations cannot be processed efficiently on conventional generations of GPUs.  
LlaMA-2 7B
\bigskip 

\emph{Conclusions.-} In this paper, we have introduced and benchmarked \emph{CompactifAI}, a compression method of Large Language Models based on quantum-inspired Tensor Networks. The method decomposes weight matrices in Self Attention and Multi-layer Perceptron layers of the LLM in terms of Matrix Product Operators with a given bond dimension, effectively truncating in the correlations present in the system. The compression rate of the model can be controlled via the bond dimension and the model accuracy can be ramped up with a short retraining (healing) process. We have shown that a combination of \emph{CompactifAI} with quantization allows to reduce a 93\% the memory size of LlaMA-2 7B, reducing also 70\% the number of parameters, accelerating 50\% the training and 25\% the inference times of the model, and just with a small accuracy drop of 2\% - 3\%. This goes much beyond of what is achievable by other compression techniques and also shows that standard LLMs are, in fact, heavily overparametrized, something that ``natural intelligence" is definitely not doing.   


Our work provides a much more refined, controllable, and explainable compression technique of LLMs compared to alternative methods such as pruning, distillation, quantization, and low-rank approximations. Furthermore, our TN method is compatible with all these techniques and can be applied alongside with them, as we have shown in this paper. This can be further explored in future works, as well as more advanced TN compression techniques for LLMs. 

In our opinion, our work opens the door to the democratization of LLMs. Small-size LLMs are a necessity to lower their gargantuan energy consumption, and can also be deployed on premises without the need of a cloud connection to somebody else's server, in turn opening an entire new world of possibilities for personalized-AI models. In our opinion, \emph{CompactifAI} and tensor network methods are going to play a fundamental role in the development of the next generation of AI technology.

\bigskip 
{\bf Acknowledgements:} We acknowledge Donostia International Physics Center (DIPC), Ikerbasque, Basque Government, Diputaci\'on de Gipuzkoa, European Innovation Council (EIC), and Spanish Government for constant support, as well as insightful discussions with the team from Multiverse Computing. A. T.  acknowledges funding by MICIN within NextGenerationEU(PRTR-C17.I1) program and by Generalitat de Catalunya. S. S. J. also acknowledges the Institute for Advanced Studies in Basic Sciences (IASBS).

\bigskip 
{\bf Data availability statement:} all data required for this project can be accessed upon reasonable request by contacting the authors. 

\bibliography{references} 

\newpage

\onecolumngrid 

\section{Supplementary Information: Layer Sensitivity Profiling of LlaMA-2 7B}

\begin{figure*}[h]
    \centering
    \includegraphics[width=\linewidth]{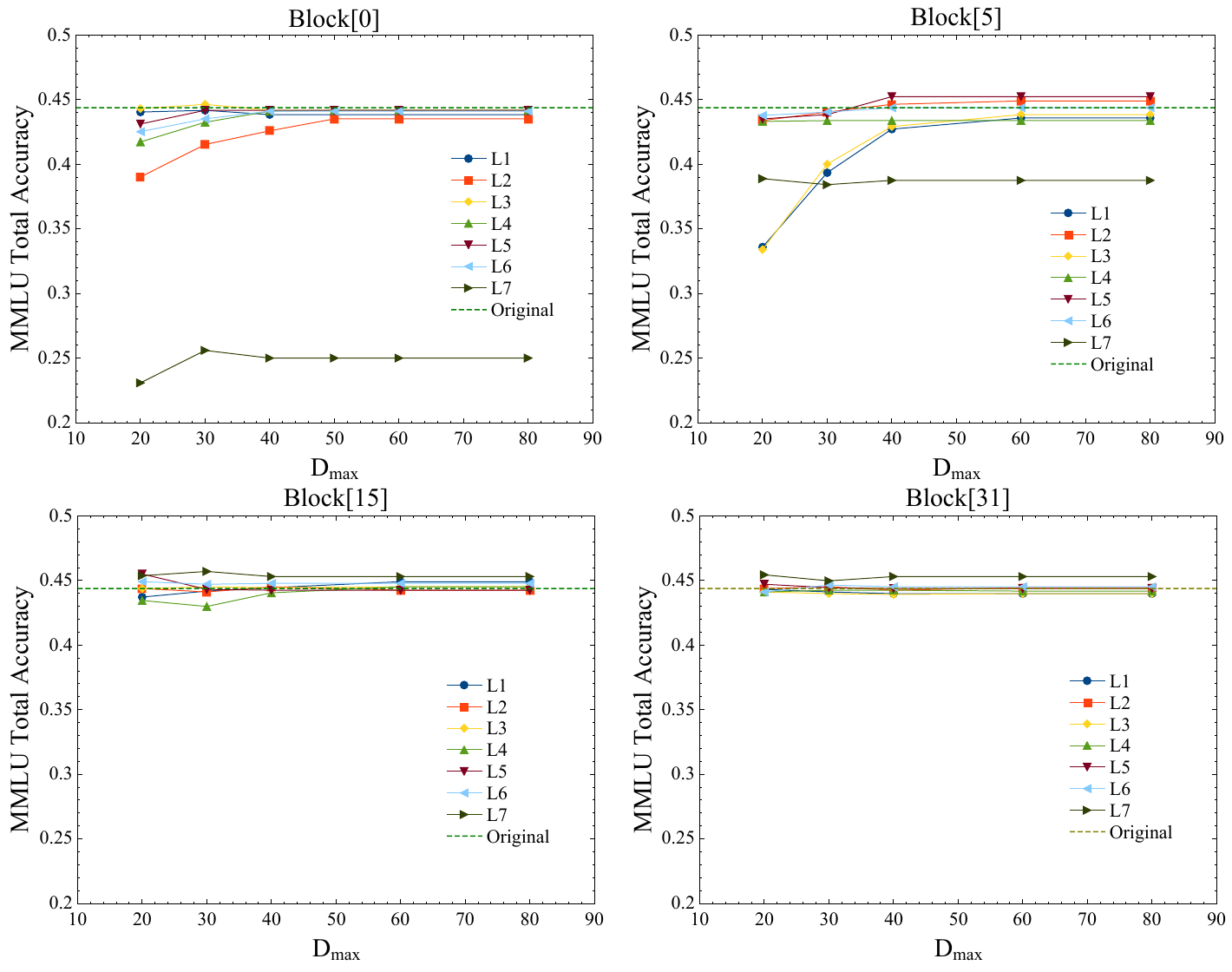}
    \caption{[Color online] Layer sensitivity analysis for layers in several attention blocks of the LlaMA-2 7B model. Layers L1-L4 correspond to the multi-head attention layers and L5-L7 are MLP layers. L7 is the last layer which also accounts for the output of the attention block. The Layers of the initial blocks are more sensitive to compression. However, the middle to the end blocks are robust to very large compressions.}
    \label{Fig:profiler}
\end{figure*}


The main building blocks of LlaMA-2 7B model are 32 attention blocks each of which is composed of four multi-head attention layers and three multi-layer perceptron layers. Depending on the location of the attention blocks, its internal layers have different sensitivity to tensor network decompositions. To have the most efficient tensor network compression, we developed a tool for profiling the layers of the LLM to assess its sensitivity to the compression level. Fig. \ref{Fig:profiler} shows the profiling results for the total MMLU accuracy versus the maximum tensor bond dimensions $D_{\rm max}$ that emerge during the tensor decomposition of layer weights. We have shown results for four different blocks from the beginning, the middle, and at the end of the LlaMA network of attention blocks, i.e., blocks 0, 5, 15, and 31. Let us further note that as the bond dimension becomes smaller, more compression happens which implies more parameters were discarded during tensor decomposition of weights. 

Fig. \ref{Fig:profiler} shows that for the LlaMA-2 7B model, the initial layers and attention blocks which are located at the beginning of the attention network are more sensitive to truncation and compression. However, as we move towards the end of the network, the sensitivity decreases and we can compress the layers down to $10\%$ of the original size without significant loss of accuracy. This, in turn, is compatible with previous observations on the ineffectiveness of deeper layers \cite{mit}. The results suggests that the attention blocks towards the middle to end of the attention networks are more suitable for large-scale tensorization and truncation. However, the layers at the beginning should be handled with more care and it is advised not to compress them below $50\%$. Another observation from Fig.\ref{Fig:profiler} is that the last MLP layer in each attention block, which amounts to the output of the block, is more sensitive to compression in the LlaMA-2 7B model. It is therefore advised that this layer be excluded from tensorization and compression in all attention blocks.   

Last but not least, we have used the MMLU accuracy as the measure of layer sensitivity of the layers. This can be done by using any dataset and metric which is relevant to the underlying model. The tensorized models in this work are all created after such a layer analysis to obtain the best-performing model. 

\end{document}